\documentclass[12pt]{article}
\usepackage[top=1in, bottom=1in, left=1in, right=1in]{geometry}
\usepackage{natbib}

\usepackage[utf8]{inputenc} 
\usepackage[T1]{fontenc}    
\usepackage{hyperref}       
\usepackage{url}            
\usepackage{booktabs}       
\usepackage{amsfonts}       
\usepackage{xfrac}       
\usepackage{microtype}      

\usepackage{xr}

\usepackage{caption} 
\usepackage{float}   

\usepackage{natbib}
\usepackage{amsthm}
\usepackage{amsfonts}
\usepackage{amsmath}
\usepackage{authblk,array,setspace} 
\usepackage{enumitem}
\usepackage{graphicx}

\theoremstyle{plain}

\newtheorem{lemma}{Lemma}

\newtheorem{definition}{Definition}

\newtheorem{remark}{Remark}
\newtheorem{step}{Step}

\newtheoremstyle{mystyle}
  {\topsep}
  {\topsep}
  {\normalfont}
  {0pt}
  {\bfseries}
  {:}
  {\newline}
  {\thmname{#1}\thmnumber{}\thmnote{ #3}}

{
\theoremstyle{mystyle}
\newtheorem{refproof}{Proof}
}
\usepackage{pgffor}%

\graphicspath{{art}}
\newcommand{\imagefolder}{art}

\makeatletter
\newcommand{\specificthanks}[1]{\@fnsymbol{#1}}
\makeatother
\title{Counterfactual Distribution Regression \\ for Structured Inference}
\author{Nicolo Colombo \thanks{\texttt{nicolo.colombo@ucl.ac.uk}}  

	\emph{ \small Department of Computer Science, 
	Royal Holloway University London}, 
\and
Ricardo Silva 
\thanks{\texttt{ricardo.silva@ucl.ac.uk}}

	\emph{\small Department of Statistical Science, 
	University College London}, 
\and
Soong M. Kang
\thanks{\texttt{smkang@ucl.ac.uk}}

	\emph{\small School of management, 
	University College London}, 
\and
Arthur Gretton
\thanks{\texttt{arthur.gretton@gmail.com}}

 \emph{\small Gatsby Computational Neuroscience Unit}, 
}

\begin{document}

\maketitle
\begin{abstract}
 We consider problems in which a system receives external
\emph{perturbations} from time to time. For instance, the system can
be a train network in which particular lines are repeatedly
disrupted without warning, having an effect on passenger
behavior. The goal is to predict changes in the behavior of the
system at particular points of interest, such as passenger traffic
around stations at the affected rails. We assume that the data
available provides records of the system functioning at its
``natural regime'' (e.g., the train network without disruptions) and
data on cases where perturbations took place. The inference problem
is how information concerning perturbations, with particular
covariates such as location and time, can be generalized to predict
the effect of novel perturbations. We approach this problem from the
point of view of a mapping from the counterfactual distribution of
the system behavior without disruptions to the
distribution of the disrupted system. A variant on
\emph{distribution regression} is developed for this setup.

\end{abstract}

\section{Introduction}

\subsection{Contribution}
Consider a complex system such as the London Underground, a large
network of fast trains for the daily commute of passengers. Measures
such as the number of passengers exiting at each station can be used
to quantify the behavior of the system. Unplanned disruptions
sometimes happen, which stop trains from running within particular
segments of the network. A local disruption has effects elsewhere in
the system, primarily in stations adjacent to the disrupted region. As
discussed by \cite{silva:15}, there is enough structure in the system
such that past disruptions can inform predictions of what will happen
under a novel disruption that takes place at a previously unseen location.

We will call a \emph{perturbation}, or shock, any kind of external
event that directly changes a particular mechanism in the system. For
instance, a signal failure in the Underground will stop trains from
navigating through particular rails. Even though the local nature of
the perturbation can be assumed to be known, with obvious immediate
effects (e.g., no trains in particular lines), a perturbation will
also have effects elsewhere in the system, which may need to be learned
from data. For instance, passengers who cannot reach a particular
destination may decide to leave at a different location. The number of
passengers exiting a station may go up or down depending on the flow
of passengers quitting earlier or not being able to reach it. Assuming
access to a historical database of past disruptions and assumptions
about invariances of particular components of the system, a model
can estimate the effect of line closures on passenger behavior,
a quantity of interest for policies that attempt to mitigate the
effect of such perturbations (such as crowd management and
compensation for excess demand by auxiliary services such as buses).

In this paper, we consider the general setting of building predictive
models for the effects of perturbations on a system. We borrow
concepts from causal inference, in particular \emph{counterfactual
  modeling}, where given a model for the ``natural regime'' of the
system, \emph{i.e.}, its usual dynamics, at the moment of a shock we
generate a distribution over counterfactual outcomes, which is a
probabilistic assessment of its possible trajectories had no shock
taken place, given the history of the system up to that moment.  We
assume the existence of a mapping from the counterfactual distribution
of a set of variables in the system to the ``factual'' distribution of
target system variables, indexed by covariates describing the
perturbation.

\subsection{Setup}
\label{sec:setup}

Consider a probability distribution $P_{\rm natural} \in \mathcal P$, where
$\mathcal P$ is a given function space, describing the \emph{natural
	regime} of an observed $D$-dimensional random variable, \emph{i.e.},
$Y = [Y_1, \dots, Y_D] \sim P_{\rm natural}$.  The distributions $P_{\rm natural} $ is defined
over an undirected graph, $\mathcal G = (\mathcal V, \mathcal
E)$, with a set of vertices $\mathcal V$ such that $|\mathcal V| = D$,
and where the edge set $\mathcal E$ forms an adjacency matrix $A \in
\{0, 1\}^{D \times D}$. Within the context of our working example, $A_{dd'} = 1$
denotes whether stations $d$ and $d'$ are physically adjacent in the
London Underground, $A_{dd'} = 0$ otherwise, with $Y_{d}$ being the
number of passengers exiting at station $d$ within a given time
window.

The natural regime can be perturbed by external events. The goal is to
predict changes induced by such external events on $Y$.  Let $\{
P_{\rm perturbed}^{(k)} \}_{k = 1}^{K}$ be the perturbed distributions associated
with $K$ such events.  We assume the $k$th perturbation to be fully
characterized by set of features $ z^{(k)}  = [d^{(k)}, u^{(k)}]
$, where $d^{(k)} \in \mathcal V$ can be interpreted as the set of
locations where the perturbation is applied to the network and
$u^{(k)}$ belongs to an arbitrary feature space $ {\cal Z} $.  In
general, a perturbation may affect all $\mathcal V$ via indirect
effects that depend on the graph structure.  In the simplest setting,
for example, one can assume a perturbation applied to $d^{(k)}$ to
have visible effects to all $d$ such that $A_{dd^{(k)}} = 1$.

Given a dataset of observations from the \emph{natural regime} 
\begin{eqnarray}
{\cal D}_{\rm natural} =  \{y^{(n)} \ {\rm realization \ of } \ Y \sim P_{\rm natural} \}_{n = 1}^{N_{\rm natural}} ,
\end{eqnarray} 
and a dataset of observations and features from different \emph{perturbed regimes}
\begin{eqnarray}
{\cal D}_{\rm perturbed} = \left\{ {\cal D}_k, z^{(k)} \right\}_{k = 1}^{K} , \qquad {\cal D}_k  =
\{ \tilde y^{(n)} \ {\rm realization \ of } \ \tilde Y \sim P_{\rm perturbed}^{(k)}\}_{n = 1}^{N_k} .
\end{eqnarray}
we describe a method where when a new perturbation
applied to node $d^{(\rm new)}$, it predicts (marginals of)
distribution $P_{\rm perturbed}^{(\rm new)} \in {\cal P}$ describing the system measurements $\tilde Y^{({\rm new})}$
under the new perturbation $z^{(\rm new)}= [d^{(\rm
	new)}, u^{(\rm new)}]$. We define perturbations as applied to single nodes, thus
allowing real-world events to generate multiple perturbation data points.

The core methodology is to cast $P_{\rm perturbed}^{(\rm new)} \in {\cal P}$ as the output of a
\emph{distribution regression model} \citep{sutherland2012kernels}, a class of regression
models where the covariates are the corresponding
natural regime distribution $P_{\rm natural}$, the perturbation features $z^{(k)} =
[d^{(k)}, u^{(k)}]$ and the adjacency matrix $A$.  More
explicitly, we seek a model ${\Psi}: {\cal P} \times {\mathcal V}
\times {\cal Z} \times \{0, 1\}^{D \times D} \to {\cal P}$, such that
we can predict the exit counts distribution under a new unseen
perturbation associated with features 
$z^{({\rm new})}= [d^{({\rm new})}, u^{({\rm new})}]$ by letting
\begin{equation}
\label{eq:psi_mapping}
P_{\rm perturbed}^{({\rm new})} = \Psi(P_{\rm natural}, x^{({\rm new})}, A),
\end{equation} 
\noindent where $P_{\rm natural} $ is a ``counterfactual'' distribution, \emph{i.e.},
the distribution of system variables $Y \sim P_{\rm natural} $ had no disruption
taken place. Unlike standard regression models, $P_{\rm natural}$ is an
unobservable structured covariate that is estimated from the
data at hand. 
This data is independent of the data used in the estimation
of (\ref{eq:psi_mapping}), as they represent different regimes.

\subsection{Relationship to Causal Modeling}
Equation (\ref{eq:psi_mapping}) is motivated by predictive causal
modeling problems where given data from the observational regime of a
system and from (possibly) a set of experiments, we infer the predictive
distribution of the outcomes of a system under an intervention.  These problems 
can be found in the machine learning literature \citep{sgs:00,pearl:00}
and in some recent advances in the statistics literature such as
\cite{peters:16}. In our approach, we do not directly target the estimation of causal
effects, such as the difference in expected outcomes between two
levels of treatment \citep{imbens:15}.  Instead, given a disruption
\citep[which can be interpreted as type of \emph{natural experiment},][]{dunning:12},
we predict what will happen to a
system in a near future. The counterfactual state of the system,
\emph{i.e.}, its probabilistic behaviour had no disruption taken place, plays
an important role, but only as a useful covariate that can aid
prediction.

The distribution of outcomes under an intervention can be inferred by
combining observational data and data from the system under
interventions.  In the context of causal graphical model
\citep{pearl:00,sgs:00}, the operational definition of an intervention
is the modification of one of the factors of the observational
distribution function, which factorizes according to a directed
\emph{causal graph}. Assuming that the density function exists, the
causal graph implies that the joint density function is given by the
product of conditional density functions $f(Y_d\ |\ Y_{\pi(d)})$,
where $Y_{\pi(d)}$ are the parents (direct causes) of $Y_d$ in the
graph \citep{pearl:00}.  An intervention $*$ on $Y_d$ changes
$f(Y_d\ |\ Y_{\pi(d)})$ into some
$f^{*}(Y_d\ |\ Y_{\pi(d)})$. Conditional densities can also be
expressed in terms of equations $Y_d = f_d(Y_{\pi_d}, \epsilon_d)$,
where $\epsilon_d$ are unobservable causes of $Y_d$. For example, an
intervention on a random variable $Y_d$ is said to be \emph{perfect}
if it replaces its natural regime equation, say $Y_d =
\alpha_{\pi_{d1}}Y_{\pi_{d1}} + \alpha_{\pi_{d2}}Y_{\pi_{d2}} +
\epsilon_d$, by the equation $Y_d = y$ for some constant $y$. This is
to be contrasted with a \emph{soft intervention}, which modifies the
equation for $Y_d$ while keeping a dependence on at least some
of its direct causes. For example, a soft intervention on
$Y_d$ that weakens its response with respect to one of its direct causes
can be modelled parametrically by a regime parameter $\beta \in
[0, 1]$, as in $Y_d = \alpha_{\pi_{d1}}Y_{\pi_{d1}} +
\alpha_{\pi_{d2}}\beta Y_{\pi_{d2}} + \epsilon_d$. A modelling choice
is to assume that this parameter $\beta$ appears on interventions at
other variables $Y_{d'} \neq Y_d$, as an example of sharing 
statistical strength between different interventions.
See \cite{sach:15} for a recent example of this approach to modelling
in applied biology, and early work by \cite{cooper:99b} and
\cite{tian2001causal} on causal inference by combining multiple
interventions and natural experiments. In particular, 
parameters for the natural regime, such as
$\alpha_{i\pi_{d1}}$, are also present in the modelling of the
interventional regime.  It is explicit here that the natural
regime informs the interventional one.

Modeling the predictive distribution of soft interventions by first
learning of a full causal graphical model requires assumptions about
how such interventions interact with unobserved causes $\epsilon_d$.
In our example above, the contribution of $\epsilon_d$ was left
unchanged. This type of modelling is particularly difficult if the
different unobserved causes are confounded, possibly requiring strong
assumptions about the parametric shape of such interventions. This is
not necessary in our setup because we are not trying to estimate a
full causal graphical model, which encodes the effect of perfect
interventions and how unmeasured confounding takes place. Instead, our
assumption is that perturbations come from a family of interventions
where invariance (\ref{eq:psi_mapping}) holds, while making no claims
concerning the representation of perfect interventions and arbitrary
causal effects.  This assumption requires that we have access to a set
of datasets collected under different perturbations, including the
natural regime, so that predictions can be made on a new unseen
perturbation that is different from the existing ones but which is
assumed to fit within the postulated relationship.

In particular, our graph $\mathcal G$ is not a causal graph in the
sense of encoding the Causal Markov condition \citep{sgs:00}, but a
symmetric graph of physical constraints as in the analysis of
\emph{interference} in social networks and spatial effects \citep[see
  e.g.][]{aronow2017estimating,aronow2017estimating2}. The graph is
assumed as part of the data, as opposed to a quantity to be
estimated. Physical-constraint graphs can be used to inform the
learning of a causal graph as done by \cite{friedman:11}, but here we
are interested solely in predicting the effect of natural experiments
coming from an unknown set of soft interventions. Equation
(\ref{eq:psi_mapping}) replaces the typical assumptions of invariances
in causal graphical modelling with a more black-box approach for
predictive modelling under regime changes. By using Equation
(\ref{eq:psi_mapping}), we assume that there is a common real-world
meaning for the elements in this class of perturbations, such as each
element is an unplanned partial line closure in the London
Underground.

Traditional counterfactual models \citep{imbens:15} postulate a joint
distribution among potential outcomes. While in principle we could
derive relationship (\ref{eq:psi_mapping}) not only from an explicit
model for soft interventions, but also from an explicit latent
variable for the joint distribution of potential outcomes, this again
requires strong assumptions, some of which are untestable since only
one potential outcome is observable at a time.  However, by exploiting
the \emph{distribution} of the counterfactuals as inputs to outcomes
of interest, as opposed to using the \emph{latent values} of the
counterfactuals, we sidestep the computational complexity of modelling
the distribution of the observed disrupted variables as the
marginal of a possibly high-dimensional missing data model.

\subsection{Relationship to Standard Distribution Regression}

Our problem is formulated as a distribution regression problem whose
theoretical and practical aspects have been largely addressed in the
literature \citep{sutherland2012kernels, lampert2015predicting,
  szabo2016learning, law2018bayesian}. To the best of our knowledge,
the definition of a perturbation map has not been considered in the
past. The perturbation map can be interpreted as a distribution
regression where both the regressor and the output are distributions,
unlike the classical setting \citep{sutherland2012kernels} where the
input are sets of observations. The goal is to estimate a mapping
between the input and output distributions.

\section{Method}
\label{sec:method}

\subsection{Distribution-to-distribution regression model}
Let ${\mathbf P}: {\cal P} \times {\cal Q} \to [0, 1]$ be a
probability distribution defined over the product function space ${\cal P}
\times {\cal Q}$.  A distribution-to-distribution regression model is
a functional $\psi : {\cal Q} \to {\cal P}$ such that
\begin{eqnarray}
\label{meta distribution}
 E_{P, Q \sim \mathbf P}(P~ |~ Q)  = \psi(Q), \qquad P \in {\cal P}, \qquad Q \in {\cal Q},
\end{eqnarray}
where $ E_{X \sim \mathbf P}(X)$ is the expectation of $X$ with respect
to the joint distribution ${\mathbf P}$. For simplicity, we
consider the case of a single input distribution. Handling multidimensional
inputs is conceptually a direct extension.


\subsection{Reproducing kernel Hilbert spaces}
We follow the framework of casting distribution regression in terms
of reproducible kernel Hilbert spaces \citep{muandet2017kernel}.
A vector space ${\cal H}$ with inner product $\langle \cdot, \cdot
\rangle_{\cal H} : {\cal H} \times {\cal H} \to {\mathbf R}$ is an
Hilbert space if it is \emph{complete} with respect to the norm $\|
\cdot \|_{\cal H}: {\cal H} \times {\cal H} \to {\mathbf R}_+ $
defined by $ \| f \|_{\cal H}^2 = \langle f, f \rangle_{\cal H} $, for
all $f \in {\cal H}$.

Given an input space ${\cal X}$, a \emph{reproducing kernel Hilbert space}
(RKHS) over ${\cal X}$ is an Hilbert space of functions $f: {\cal X}
\to {\mathbf R}$ that satisfy the additional smoothness condition 
\begin{eqnarray}
\ \sup_{x \in \cal X} f(x) \leq C \| f \|_{\cal H}, \quad C < \infty . 
\end{eqnarray}  
A RKHS ${\cal H}$ is completely characterized by its reproducing
kernel, a symmetric and positive-definite function $k :
{\cal X} \times {\cal X} \to {\mathbf R}_+$ that satisfies the
\emph{reproducing property},
\begin{eqnarray}
f(x) = \langle f, k(x, \cdot) \rangle_{\cal H}, \qquad {\rm for \ all}
\ f \in {\cal H} \ {\rm and \ all } \ x \in {\cal X}.
\end{eqnarray} 
This implies $ k(x, x') = \langle k(x,\cdot), k(x', \cdot)
\rangle_{\cal H}  =  \langle \phi(x), \phi(x') \rangle_{\cal H}$,
where $\phi(x)  =  k(x, \cdot)$ is usually referred to as the
\emph{canonical feature map} $\phi(\cdot): {\cal X} \to {\cal H}$.

\subsection{RKHS embedding of distributions}

The canonical feature map can be used to obtain an Hilbert space
representation of any element of the input space $x \in {\cal X}$.
Let $P:{\cal X} \to [0, 1]$ be a probability distribution defined over
${\cal X}$. Its Hibert space representation is then defined by
\begin{eqnarray}
\mu_P  = E_{X \sim P}(k(X, \cdot)) \in {\cal H},
\end{eqnarray} 
and it is called the \emph{RKHS mean embedding} of $P$.
Under mild conditions on $k$, $\mu_P$ is such that (\cite{smola2007hilbert})
\begin{eqnarray}
 E_{X \sim P} (f(X)) = \langle \mu_P, f \rangle_{\cal H} \qquad {\rm for \ all} \ f \in {\cal H}.
\end{eqnarray} 
If $k$ is a \emph{characteristic kernel}, the map $P \to \mu_P$ is injective and 
\begin{eqnarray}
\|\mu_P - \mu_Q \|_{\cal H} = 0 \quad {\rm if \ and \ only \ if }  \quad P = Q.
\end{eqnarray} 
For example, both the Laplace and Gaussian kernels defined by $k(x,
x') = \exp{ (- \rho \| x - x' \|_1)}$ and $k(x, x') = \exp{ (- \rho \|
	x - x' \|^2)}$ are characteristic on ${\mathbf R}^N$.  More
generally, a kernel is characteristic if it is translation invariant
and its Fourier transform has support over the whole space
(see \cite{sriperumbudur2010hilbert} and also \cite{fukumizu2004dimensionality} for more details
on the definition of the class of characteristic kernels.).  For any
distribution $P$, an empirical estimation of $\mu_P$ is
\begin{eqnarray}
\label{empirical mean embedding}
\hat \mu_{P} = |{\cal D}_{P}|^{-1} \sum_{x \in {\cal D}_{P}} k(x,
\cdot) \in {\cal H}, \qquad {\cal D}_{P} = \{ x^{(n)} \in {\cal X}
\ {\rm realization \ of } \ P \}_{n = 1}^{N_P}.
\end{eqnarray} 
It can be shown that $\hat \mu_P$ is an unbiased estimate of $\mu_P$
\citep{sriperumbudur2012empirical} and 
\begin{eqnarray}
\|  \mu_P - \hat \mu_P \|_{\cal H}  = O\left( |{\cal D}_P|^{-1} \right).
\end{eqnarray}

\subsection{Regression in the Hilbert space}
Given a distribution-to-distribution regression model,
$\Psi: {\cal Q} \to {\cal P}$, and a kernel function that is characteristic on
${\cal X}$, the  RKHS regression
model, $L_{\Psi} \in {\cal L}_{{\cal Q}{\cal P}}$, is a linear
operator from the RKHS of ${\cal P}$ to the RKHS of ${\cal Q}$.

\begin{definition}[Non-parametric model]
The non-parametric model is a linear operator $L_{\Psi}: {\cal H}_{Q} \to {\cal H}_{P}$ implicitly defined by $L_{\Psi} \ \mu_{P^{(k)}} =
\mu_{\psi(Q^{(k)})}$, where $\{[P^{(k)}, Q^{(k)}] \}_{k=1}^K$ are realizations
of ${\mathbf P}$.	
\end{definition} 


Given a dataset of samples drawn from a set of input-output distributions 
\begin{eqnarray}
\label{datasets}
{\cal D}^{(K)} = \{ {\cal D}_{P^{(k)}}, {\cal D}_{Q^{(k)}}\}_{k = 1}^K,
\quad {\cal D}_{U} = \{ u^{(n)} \ {\rm realization \ of \ } U\}_{n =
	1}^{N_U}, \quad U \in \{P^{(k)}, Q^{(k)}\}_{k = 1}^{K}, \quad  
\end{eqnarray}
let $\hat M_U = [\hat \mu_{U^{(1)}}, \dots, \hat \mu_{U^{(K)}}]$ ($U \in \{P, Q\}$).
Then the optimal RKHS regression model is 
\begin{eqnarray}
\label{estimator}
\hat L_{\Psi}^{(K)} = {\rm arg} \min_{L \in {\cal L}_{{\cal Q}{\cal P}}} \| \hat M_{P} - L \hat M_{Q} \|^2_{\cal H}  
= \hat M_P \hat M_Q^T (\hat M_Q \hat M_Q^T)^{-1} 
=  \hat M_P (\hat M_Q^T  \hat M_Q)^{-1} \hat M_Q^T . 
\end{eqnarray}
The empirical mean embeddings $\{\hat \mu_{P^{(k)}}, \hat \mu_{Q^{(k)}} \}_{k = 1}^K$
are obtained from \eqref{datasets} via \eqref{empirical mean embedding}.  Note that, for
finite $K$, the operator acts non-trivially only on the
$K$-dimensional subspace spanned by the columns of $\hat M_Q$, which
is assumed to have full column rank.

\begin{lemma}[Estimation of the non-parametric model]
	\label{lemma consistency non parametric}
	Assume that the meta-distribution ${\mathbf P}: {\cal P} \times {\cal
		Q} \to [0, 1]$ generating the realization dataset ${\cal D}^{(K)}$ defined in \eqref{datasets} can be
	represented by a \emph{unique} RKHS linear operator ${\cal L}_{{\cal
			Q}{\cal P}}$ such that
	\begin{eqnarray}
	\label{exact model non parametric}
	\mu_{P} = L_{\rm true} \mu_{Q}, 
	\end{eqnarray}
	for any $[P, Q] \sim
	{\mathbf P}$.
	Then the estimator \eqref{estimator} restricted to the subspaces spanned by the columns of $\hat M_U = [\hat \mu_{U^{(1)}}, \dots, \hat \mu_{U^{(K)}}]$ ($U \in \{P, Q\}$) is consistent, \emph{i.e.},
	\begin{eqnarray}
	\left\| {\hat \Pi}_{P}^{(K)} L_{\rm true} {\hat \Pi}_{Q}^{(K)} - {\hat \Pi}_{P}^{(K)} \hat L^{(K)} {\hat \Pi}_{Q}^{(K)} \right\| = O(1/ D^{(K)}),  
	\end{eqnarray}
	where $\hat \Pi_{U}^{(K)}  = \hat M_U (\hat M_U^T \hat M_U)^{-1} \hat M_U^T$ ($
	U \in \{P, Q\}$) and 
	$D^{(K)} = \min\{ \min\{ |{\cal D}_{P^{(k)}}|, |{\cal D}_{Q^{(k)}}|\}$ ($ {\cal D}_{P^{(k)}}, {\cal D}_{Q^{(k)}} \in {\cal D}^{(K)} $).
\end{lemma}

\subsection{Finite-dimensional parametrizations of $L_{\psi}: {\cal H}_{\cal Q} \to {\cal H}_{\cal P}$}
\label{section mixture of mean embeddings}
In some cases, as in the experiments shown in Section
	\ref{sec:experiments}, one may prefer to choose a
finite-dimensional parametrization of $L_{\Psi}$.  Here, we give two
examples where structural assumptions are made directly on the RKHS
linear operator $L_{\Psi}: {\cal H}_{\cal Q} \to {\cal H}_{\cal P}$.
The alternative approach where structural constraints are imposed
directly on the distribution-to-distribution regression model, $\Psi
:{\cal Q} \to {\cal P}$, is in general harder and will be considered
in Section \ref{section mixture of distributions}. Here, we assume for
simplicity that all $Q^{(k)}$ and $P^{(k)}$ belong to the same
function space, \emph{i.e.}, where ${\cal P} = {\cal Q}$ and ${\cal
	H}_{\cal P} = {\cal H} = {\cal H}_{\cal Q}$ and focus on two specific models:
a  \emph{one-parameter model}, where the RKHS operator $L_{\Psi} :
{\cal H} \to {\cal H}$ is defined by $L_{\Psi} f = \alpha f$ for all
$ f \in {\cal H}$ and $\alpha \in {\mathbf R}$ and a \emph{mixture of mean embeddings}, where the  RKHS
operator $L_{\Psi} : {\cal H}^{\otimes I} \to {\cal H}$ is defined by
$L_{\Psi} [f_1, \dots, f_{I}] = \sum_{i = 1}^I \alpha_i f_i$, for all
$f_i \in {\cal H}$, $i = 1, \dots, I$.

\begin{definition}[One-parameter model]
\label{definition one parameter model}
The one-parameter model is a linear operator
$L_{\Psi} : {\cal H}\to {\cal H}$ defined by 
\begin{eqnarray}
\label{one parameter model}
L_{\Psi} = \alpha 1_{\cal H}, \qquad \alpha \in {\mathbf R}, 
\end{eqnarray} 
where the RKHS identity operator $1_{\cal H} \in {\cal L}_{{\cal P}
	{\cal P}}$ is defined by $f = 1_{\cal H} f$ for any $ f \in {\cal
	H}$.  
\end{definition}

This is the simplest
possible non-trivial linear operator in $L_{\Psi} : {\cal H} \to
{\cal H}$. 
Given the training sample, ${\cal D}^{(K)}$, a least-squares
estimate of the free parameter is
\begin{eqnarray}
\label{estimator one parameter model}
\hat \alpha^{(K)} = \left({\rm trace}(\hat m_{QQ} \right)^{-1}
\left({\rm trace}(\hat m_{PQ} )\right) \quad 
 \left[\hat m_{UV}
\right]_{kk'} = \left[\hat M_U^T \hat M_{V}\right]_{kk'} \quad  U, V \in \{P, Q \}, 
 \end{eqnarray}
where $ [\hat m_{UV}]_{kk'} = \left(|{\cal D}_{U^{(k)}}| |{\cal D}_{V^{(k')}}|\right)^{-1} \sum_{u \in
	{\cal D}_{U^{(k)}}} \sum_{v \in {\cal D}_{V^{(k')}}} k(u, v)$, for all $k= 1, \dots, N_k$ and $k'=1, \dots, N_{k'}$.

\begin{lemma}[Estimation of the one-parameter model]
	\label{lemma consistency of the one-parameter model}
	Assume that the meta-distribution ${\mathbf P}: {\cal P} \times {\cal
		Q} \to [0, 1]$ generating the dataset ${\cal D}^{(K)}$ defined in \eqref{datasets} can be
	represented by a \emph{unique} RKHS linear operator ${\cal L}_{{\cal
			Q}{\cal P}}$ such that
\begin{eqnarray}
 \mu_{P} = \bar \alpha \mu_{Q}, 
\end{eqnarray} 
for any $[P, Q] \sim {\mathbf P}$.
Then the estimator \eqref{estimator one parameter model} obeys
\begin{eqnarray}
|\hat \alpha^{(K)}  - \bar \alpha |  =  O\left( 1/D^{(K)}\right),  
\end{eqnarray}
where $D^{(K)} = \min\{ \min\{ |{\cal D}_{P^{(k)}}|, |{\cal D}_{Q^{(k)}}|\}$ ($ {\cal D}_{P^{(k)}}, {\cal D}_{Q^{(k)}} \in {\cal D}^{(K)} $).
\end{lemma}

\begin{definition}[Mixture of embeddings]
\label{definition mixture of embeddings}
The mixture of embeddings model is a linear operator $L_{\Psi} : {\cal H}^{\otimes I} \to {\cal H}$ defined by 
\begin{eqnarray}
\label{mixture of embeddings}
L_{\Psi} [f_1, \dots, f_I]  =  [f_1, \dots, f_I] \alpha = \sum_{i=1}^I f_i \alpha_i,
\quad \alpha \in {\mathbf R}^I, 
\end{eqnarray}
where $f_i \in {\cal H}$ ($i = 1, \dots, I$).  
\end{definition} 
Given a meta-distribution ${\mathbf P}: {\cal P}^{\otimes
	I} \times {\cal P} \to [0, 1]$ we consider the dataset
\begin{eqnarray}
\label{dataset multiple input}
{\cal D}^{(K)}  &=& \{  {\cal D}_{U^{(k)}}, U^{(k)} \ {\rm realization \ of } \ [P, \{Q_i \}_{i = 1}^I]  \sim {\mathbf P} \}_{k = 1}^K \nonumber, \\
{\cal D}_{U^{(k)}}  &=& \{ [y^{(n)}, \{x^{(n)}_i \}_{i = 1}^I] \ {\rm realization \ of \ } [Y^{(n)}, \{X^{(n)}_i \}_{i = 1}^I ] \sim U^{(k)} \}_{n = 1}^{N_{U^{(k)}}},
\end{eqnarray}  
the least-squares estimate of $\alpha$ is
\begin{eqnarray}
\label{estimator mixture of embeddings}
\hat \alpha^{(K)} = \left( \sum_{k = 1}^K (\hat M_Q^{(k)})^T \hat M_Q^{(k)} \right)^{-1} \sum_{k = 1}^K (\hat M_Q^{(k)})^T \hat \mu_{P^{(k)}}, \qquad \hat M^{(k)}_Q = [\hat \mu_{Q_1^{(k)}}, \dots, \hat \mu_{Q_I^{(k)}}],
\end{eqnarray}
where the empirical mean embeddings are obtained from ${\cal D}^{(K)}$ and we assume that $K$ is big enough for the matrix $ \hat m^{(K)} = \sum_{k = 1}^K (\hat M_Q^{(k)})^T \hat M_Q^{(k)} $ to be full rank.

\begin{lemma}[Estimation of the mixture of embedding model]
	\label{lemma consistency of the mixture of embedding model}
	Assume that the meta-distribution ${\mathbf P}: {\cal P}^{\otimes
		I} \times {\cal P} \to [0, 1]$ generating the realizations dataset 
	${\cal D}^{(K)}$ defined in \eqref{dataset multiple input} can be
	represented by a \emph{unique} RKHS linear operator ${\cal L}_{{\cal
			Q}{\cal P}}$ such that
	\begin{eqnarray}
	\mu_{P} = M_Q \bar \alpha = \sum_{i=1}^I \mu_{Q_i} \bar \alpha_i ,
	\end{eqnarray} 
	for any $[P, \{Q_i \}_{i = 1}^I] \sim {\mathbf P}$.
	Then the estimator \eqref{estimator mixture of embeddings} obeys
	\begin{eqnarray}
	\| \bar \alpha  - \hat \alpha^{(K)} \|  =  O\left( 1/D^{(K)}\right) , 
	\end{eqnarray}
	where $D^{(K)} = \min \{ |{\cal D}_{U^{(k)}}|, {\cal D}_{U^{(k)}} \in {\cal D}^{(K)}\}_{k = 1}^K$.
\end{lemma}

\begin{remark}
	To simplify the notation we assume that an equal
	number of samples from $P^{(k)}$ and each $Q_i^{(k)}$ ($k = 1, \dots,
	K$, $i=1, \dots, I)$ is available. 
\end{remark}

\begin{remark}
\label{remark choice for modelling}
	The mixture of embeddings model defined in Definition~\ref{definition mixture of embeddings} is an example of regression model, $\Psi : {\cal P}^{\otimes I} \to
	{\cal P}$ and $L_{\Psi}:{\cal H}^{I} \to {\cal H}$, which
	takes multiple inputs.  
	This is the model class we have used in the application
	described in Section \ref{section implementation} and corresponds
	to	a distribution-to-distribution regression model $\Psi: {\cal
		P}^{\otimes I} \to {\cal P}$, \emph{i.e.}, $ E_{P| \{ Q_i \}_{i =
			1}^I }(P) = \Psi(\{ Q_i \}_{i = 1}^I)$, where the expectation
	is over the meta-distribution ${\mathbf P}: {\cal P}^{\otimes
		I} \times {\cal P} \to [0, 1]$ generating dataset  
	${\cal D}^{(K)}$ defined in \eqref{dataset multiple input}.
\end{remark}

\begin{remark}
	Even for the simple RKHS linear operators defined in this section, it is not
	straightforward to obtain an explicit form of the corresponding
	distribution-to-distribution regression model, $\Psi :{\cal Q} \to
	{\cal P}$, which is defined implicitly by $L_{\Psi} \mu_{Q} = \mu_{\Psi(Q)}$.
	Even when all $Q_i$ have a density, it cannot be guaranteed that the
	outputs of such a model, $P^{(k)} = \Psi(Q^{(k)})$, also have a
	density.
\end{remark}

\subsection{Direct parametrization of $\Psi: {\cal Q} \to {\cal P}$}
\label{section mixture of distributions}
A more intuitive way of defining a model class is to impose a
structure directly on the distribution-to-distribution functional,
$\Psi$.  The drawback of this approach is the need to find the
structure of the corresponding RKHS linear operator $L_{\Psi}$.  This
is in general a highly non-trivial task as the constraints on
$L_{\Psi}$ are expressed in equations involving expectations of the
kernel function with respect to the input and output distributions.
More concretely, given $\Psi: {\cal Q} \to {\cal P}$, the task is to
find $L_{\Psi}$ such that
\begin{eqnarray}
 E_{[P, Q] \sim {\mathbf P}} \left( L_{\Psi} E_{X \sim Q}(k(X, \cdot)) \right) =
 E_{[P, Q] \sim {\mathbf P}} \left(E_{X \sim \Psi(Q)}(k(X, \cdot)) \right).
\end{eqnarray}

A special case where a possible parametric version of $L_{\Psi}$ can
be obtained directly from the corresponding parametric version of
$\Psi$ is when $\Psi$ is a mixture of distributions, \emph{i.e.},
\begin{eqnarray}
\Psi(\{Q_i\}_{i = 1}^I) = \sum_{i=1}^I w_i Q_i, \qquad w\geq 0, \qquad 1^Tw = 1.
\end{eqnarray}
In this case, we need to solve 
\begin{eqnarray}
\label{hilbert operator mixture of distributions equation}
 E_{[P, \{Q_i\}_{i = 1}^I] \sim {\mathbf P}}
\left( L_{\Psi} [ \mu_{Q_1}, \dots, \mu_{Q_I}]  -
\sum_{i =1}^I w_i \mu_{Q_i} \right) = 0, 
\end{eqnarray}
where $\mu_{Q_i} = E_{X \sim Q_i}(k(X, \cdot))$.
A possible solution is a linear operator as defined below.

\begin{definition}[Mixture of distributions model]
	\label{definition mixture of distributions}
	The mixture of distributions model is a linear operator $L_{\Psi} : {\rm vec}{\cal H}^{\otimes I} \to {\cal H}$, ${\rm vec}({\cal H}^{\otimes I}) = \{
	[f_1^T, \dots, f_I^T]^T, \ f_i \in {\cal H} (i = 1, \dots, I) \}$, defined by 
	\begin{eqnarray}
	\label{hilbert operator mixture of distributions}
	L_{\Psi} =  w^T \otimes 1_{\cal H},
	\quad w \geq 0, \quad 1^T w = 1,
	\end{eqnarray}
	where  $1_{\cal H}$ is the identity operator defined in Definition \ref{definition one parameter model}.  
\end{definition} 

Given a meta-distribution ${\mathbf P}: {\cal P}^{\otimes
	I} \times {\cal P} \to [0, 1]$ and the corresponding realization dataset ${\cal D}^{(K)}$ 
obtained as in \eqref{dataset multiple input}, an estimate of $w$ defined in Definition \ref{definition mixture of distributions} is 
\begin{eqnarray}
\label{estimator mixture of distributions}
\hat w = {\rm arg} \min_{w \in {\cal S}_I} \sum_{k=1}^K \| \hat \mu_{P^{(k)}} - \sum_{i=1}^I w_i \hat \mu_{Q_i^{(k)}} \|^2_{{\cal H}} ,  \quad {\cal S}_I = \{ w \in [0, 1]^I, 1^T w = 1 \},
\end{eqnarray}
where the empirical mean embeddings are obtained from ${\cal D}^{(K)}$.

\begin{lemma}[Estimation of the mixture of distributions model]
	\label{lemma consistency of the mixture of distributions model}
	Assume that the meta-distribution ${\mathbf P}: {\cal P}^{\otimes
		I} \times {\cal P} \to [0, 1]$ generating dataset 
	${\cal D}^{(K)}$ as in \eqref{dataset multiple input} is such that
	\begin{eqnarray}
	P = \sum_{i=1}^I \mu_{Q_i} \bar w_i, \quad \bar w \geq 0, \quad 1^T \bar w = 1, 
	\end{eqnarray} 
	for any $[P, \{Q_i \}_{i = 1}^I] \sim {\mathbf P}$.
	Then the estimator \eqref{estimator mixture of distributions} obeys
	\begin{eqnarray}
	\| \bar w  - \hat w^{(K)} \|  =  O\left( 1/D^{(K)}\right), 
	\end{eqnarray}
	where $D^{(K)} = \min \{ |{\cal D}_{U^{(k)}}|, {\cal D}_{U^{(k)}} \in {\cal D}^{(K)}\}_{k = 1}^K$.
\end{lemma}

\subsection{Sampling from the mean embedding}
\label{section sampling procedure}
Samples from the mean embedding of a distribution are often obtained
via \emph{herding}, which requires to solve a non-convex optimization
problem for each new sample.  It is known that herding becomes
expensive and unreliable in high dimensions. Here, we propose an
alternative method that only requires solving a single
simplex-constrained convex minimization.

Suppose we are given the empirical mean embedding, $\hat \mu \in
{\cal H}_{\cal P}$, of an unknown distribution, $P_{\hat \mu} \in
{\cal P}$, but have no access to its samples.  In the
distribution-to-distribution settings described here, $\hat \mu$ is
the output of the RKHS regression model, \emph{i.e.}, $\hat \mu = \hat
L_{\Psi} \hat \mu_{\rm input}$, but what follows may apply to more
general cases.  The task is to reconstruct the unavailable samples of
$P_{\hat \mu}$.

The strategy is to choose a basis for the target functions space
${\cal P}$ and to approximate the target distribution $P_{\hat \mu} \in
{\cal P}$ with an \emph{approximating mixture} of elements from such a
basis.
The resulting empirical mean embedding is then used to compute
an estimate of the mixture's coefficients by solving a regression
problem in the RKHS of ${\cal P}$.
This is possible
because the parameters of the mixture of distributions and the
corresponding mixture of embeddings can be chosen to be the same, as
we have shown in Section \ref{section mixture of distributions}.
Finally, approximate samples from $P_{\hat \mu}$ are obtained by
sampling from the obtained approximating mixture.

\begin{lemma}[Mean embedding sampling]
	\label{lemma mean embedding sampling}
	Let $\hat \mu$ be a given empirical mean embedding and $P_{\hat \mu} \in {\cal P}$ 
	the unknown distribution associated with $\hat \mu$. 
	Let $\{ P_i \in {\cal P} \}_{i = 1}^I$ be a suitable
	finite-dimensional basis of the functions space ${\cal P}$ and assume
	it is possible to sample from all $P_i$ ($i=1, \dots, I$).
	Then approximate samples from $P_{\hat \mu}$ are the realizations of
	\begin{eqnarray}
	\label{approximating mixture}
	X \sim \sum_{i=1}^I \hat \theta_i P_i, \quad \hat \theta = {\rm arg} \min_{\theta \in {\cal S}_I}  \| \hat \mu - \sum_{ i = 1}^I \theta_i \hat \mu_i \|^2_{\cal H}, 
	\end{eqnarray} 
	where ${\cal S}$ is defined in \eqref{estimator mixture of distributions} and $\hat \mu_i$ 
	is the empirical embedding computed from the samples of $P_i$ ($i=1, \dots, I$). 
\end{lemma}

\begin{lemma}[Consistency of the sampling scheme]
\label{lemma consistency of the sampling scheme}
Let ${\cal D}_{P_{\hat \mu}}$ be a set of samples from $P_{\hat \mu}$ (usually unavailable) and ${\cal D}_{P_{\hat \theta}}$ a dataset of samples from the approximating mixture $P_{\hat \theta}$ defined in 
Lemma \ref{lemma mean embedding sampling}.
Then 
\begin{eqnarray}
\left| \hat E_{{\cal D}_{P_{\hat \mu}}}(f) - \hat E_{{\cal D}_{P_{\hat \theta}}}(f) \right| 
 = O\left(\epsilon \right) + O\left(|{\cal D}_{P_{\hat \mu}} |^{-1}\right) + O\left(|{\cal D}_{P_{\hat \theta}}|^{-1}\right), 
\end{eqnarray}
where $\hat E_{{\cal D}_{U}} (f) = |{\cal D}_{U} |^{-1} \sum_{x \in {\cal D}_{U}} f(x) $, $\epsilon= \min_{\theta \in {\cal S}_I} {\rm dist}(P_{\hat \mu}, \sum_{i = 1}^I \theta_i P_i)$ and ${\rm dist}(P, Q)$ is a measure of the distance between $P$ and $Q$.
\end{lemma}

\section{Implementation and experiments}
\label{sec:experiments}

\subsection{Spatial interference via distribution regression}
\label{section implementation}
The general scheme proposed in Section \ref{sec:method} can be 
straightforwardly adapted to tackle the causal interference task outlined in
Section \ref{sec:setup}.
Let $A$ be the adjacency matrix of a graph ${\cal G}$ with $D$ nodes
and $Y \in {\mathcal R}^{D}$ an observed random variable defined on
${\cal G}$.  The task is to model the distribution of $Y$ under
perturbations applied at given nodes of ${\cal G}$.  For simplicity,
perturbations are assumed to be fully characterized by their ``centre
node'', $d_*$, but dependencies on extra features and possible extended
``centre regions'' can be included without major changes.  We seek a
distribution-to-distribution regression model, $\Psi$, whose input are
the adjacency matrix of the graph, $A$, the perturbation centre,
$d_*$, and the distribution of $Y$ when no perturbations are active,
$P_{\rm natural}: {\mathbf R}^D \to [0, 1]$.  We also assume that
$P_{\rm natural}$ and the target distribution $P_{\rm perturbed}:
{\mathbf R}^D \to [0, 1]$ belong to the same function space, ${\cal
	P}$.  Because of the added dependence on the graph structure we call
$\Psi$ a spatial interference distribution-to-distribution model (see Section 
\ref{sec:setup}) The underlying assumption is the existence of a
meta-distribution, ${\mathbf P}$ such that $[P_{\rm perturbed},
P_{\rm normal},d_*, A] \sim {\mathbf P}$ and 
\begin{eqnarray}
E_{P_{\rm perturbed},  P_{\rm normal},d_*, A \sim {\mathbf P}}(P_{\rm perturbed}~|~P_{\rm normal}, d_*, A) = \Psi(P_{\rm natural}, d_*, A) , 
\end{eqnarray} 
where $P_{\rm perturbed}, P_{\rm natural} \in {\cal P}$, $d_* \in
\{1, \dots, D\}$, $A \in \{0, 1\}^{D \times D}$.
The regression model $\Psi$ is then a map from ${\cal
	P} \times \{1, \dots, D\} \times \{0, 1\}^{D \times D} $ to $ {\cal
	P}$.  The task is to learn $\Psi$ given the datasets ${\cal D}_{P_{\rm natural}} = \{ x^{(n)} \ {\rm
	realization \ of } \ X \sim P_{\rm natural} \}_{n = 1}^N$, and $\{ {\cal D}_{P_{\rm perturbed}^{(k)}}, d_*^{(k)} \}_{k = 1}^K$, where ${\cal D}_U =
\{x^{(n)} {\rm realization \ of \ } U \}_{n = 1}^{N_U}$ ($U \in \{ P_{\rm perturbed}^{(k)} \}_{k=1}^K$).

\begin{step}[Preprocessing]
Reformulate the problem to frame it into the mixture of embeddings regression tasks 
described in Section \ref{section mixture of mean embeddings}. 
For each perturbation, we form a set of input distributions that
depend on $P_{\rm natural}$ and the distribution features, \emph{i.e.},
\begin{eqnarray}
Q_{i}^{(k)} = {\cal F}_i(P_{\rm natural}, z^{(k)}, A), \quad (i = 1, \dots, I, \  k = 1, \dots, K).
\end{eqnarray}
Since we only need a dataset of samples from the input distributions,
the functionals ${\cal F}_i$ are not required to have an explicit form. Let
$\{\{Q_{i}^{(k)} \}_{k=1}^K\}_{i=1}^I$  be the distributions describing a set of random variables, $\{\{ X_i^{(k)}\}_{k=1}^{K}\}_{i=1}^I$, 
that depend deterministically on 
$Y \sim P_{\rm natural}$ and the perturbation features.  For example, we can
let $\{\{ X_i^{(k)}\}_{k=1}^{K}\}_{i=1}^I$  be defined by 
\begin{eqnarray}
\label{input distributions example}
\left[ X^{(k)}_{i} \right]_d = e^{ - i \ \beta \ {\rm dist}_{A}(d, d^{(k)}_*)} [Y]_d,
\quad Y \sim P_{\rm natural},
\quad (i = 1, \dots, I, \ k = 1, \dots, K), 
\end{eqnarray}
where $\beta > 0$ and ${\rm dist}_{A}(d, d')$ is the length of the
path connecting $d$ and $d'$ on ${\cal G}$.  While figuring out a
functional form of $\{Q_{i}^{(k)} \}_{k=1}^K$ is generally difficult, we can easily
compute the corresponding empirical mean embeddings, $\{\hat
\mu_{Q_{i}^{(k)}} \}_{k=1}^K$, as in \eqref{empirical mean embedding} given the realization datasets 
${\cal D}_{U} = \{ x^{(n)} \ {\rm realization \ of \ }
X^{(k)}_{i} \sim U \}_{n = 1}^{N}$ ($U \in \{Q_{i}^{(k)},  \}_{k=1}^K$, $N
= |{\cal D}_{P_{\rm natural}}|$) obtained deterministically from ${\cal D}_{P_{\rm natural}}$ according to 
\eqref{input distributions example}. 
The mean embedding of the observed
output distributions to be used for training are computed as in \eqref{empirical mean embedding}  from $\{{\cal
	D}_{P_{\rm perturbed}^{(k)}} \}_{k = 1}^K$.
\end{step}

\begin{step}[RKHS inference]
Learn a model that relates the input embeddings, $\{\{ \hat \mu_{P_{\rm perturbed}^{(k)}}\}_{i = 1}^I\}_{k = 1}^K$, to the
outputs, $\{\{ \hat \mu_{Q_i^{(k)}} \}_{i = 1}^I\}_{k = 1}^K$.
As mentioned in Remark \ref{remark choice for modelling} of Section \ref{section mixture of mean embeddings}, we choose 
a mixture of embeddings model (see Definition \ref{definition mixture of embeddings}) with estimated parameter  \eqref{estimator mixture of embeddings}. 
Then use the trained model to compute the mean embedding of $P_{\rm
	perturbed}^{({\rm new})}$, which is the distribution associated with a new
unseen perturbation centered in $d_*^{({\rm new})}$.  
To make the prediction,
we first need to form the sample datasets corresponding to the input distributions,
$\{Q_{i}^{({\rm new})}\}_{i=1}^I$, associated with the centre of the new disruption, $d_*^{({\rm new})}$, 
and computed from $P_{\rm natural}$ as in \eqref{input distributions example}.
From such sample datasets we can compute the empirical embeddings of $\{Q_{i}^{({\rm new})}\}_{i=1}^I$
and let
\begin{eqnarray}
\hat \mu_{P_{\rm perturbed}^{(\rm new)}} =  L_{\hat \alpha} \hat
M_{Q^{({\rm new})}} = \sum_{i = 1}^I \hat \alpha_i \hat
\mu_{Q_{i}^{\rm new}} \quad \hat M_{Q^{({\rm new})}} = [\hat \mu_{Q_{1}^{\rm new}}, \dots, \hat \mu_{Q_{I}^{\rm new}} ].
\end{eqnarray}
\end{step}

\begin{step}[Sampling from the model's output]
	In order to compute approximate samples from $P_{\rm perturbed}^{(\rm new)}$, define a basis for the function space of the target distribution, ${\cal P}$.
A possible simple choice is to use the marginals of the input
distributions 
\begin{eqnarray}
\{ \{ U_{id} \in {\cal P} \ {\rm s.t.} \ X'_{id} \sim U_{id},   \ X'_{id} = {\bf e}_{d}{\bf e}_{d}^T X_i, \ X_i \sim Q^{({\rm new})}_{i} \}_{d = 1}^D \}_{i = 1}^I, 
\end{eqnarray} 
where the canonical basis vectors ${\bf e}_d \in \{0, 1 \}^{D}$ are defined by $[{\bf e}_d ]_{d'} = \delta_{d, d'} $ ($d, d'=1, \dots, D$), with $\delta_{d,d'}$ being the Kronecker delta.
The empirical embeddings of the function space basis  $\{ \{ U_{id} \}_{d = 1}^D \}_{i = 1}^I$ are obtained from 
$\{ {\cal D}_{Q^{({\rm new})}_{i}} \}_{i=1}^I$, which in
turn are obtained from ${\cal D}_{P_{\rm natural}}$.
More precisely, we compute the datasets $\{ \{ {\cal D}_{U_{id}}\}_{i=1}^I \}_{d=1}^D $ defined by 
\begin{eqnarray}
{\cal D}_{U_{id}} = \{ {\bf e}_i {\bf e}_i^T x^{(n)},  \ [x^{(n)}]_d = \ e^{ - i \ \beta \ {\rm dist}_{A}(d, d^{({\rm new})}_*)} [y^{(n)}]_d,   y^{(n)} \in {\cal D}_{P_{\rm natural}} \}_{n = 1}^{N_{\rm natural}},  \nonumber 
\end{eqnarray}
and compute the corresponding $\hat \mu_{id}$ ($i=1, \dots, I$, $d = 1, \dots, D$) as in  \eqref{empirical mean embedding}.
Then, define an approximating mixture $P_{\theta} = \sum_{i = 1}^I \sum_{d = 1}^D \theta_{id} U_{id}$ ($\theta \in [0, 1]^{ID}$, $\sum_{i=1}^I \sum_{d=1}^D \theta_{id} = 1$) of $P^{(\rm new)}_{\rm perturbed}$ and estimate the corresponding mixture weights as described in Section \ref{section sampling procedure}.
Approximated samples from $P_{\rm perturbed}^{(\rm new)}$ are then the realizations of $X \sim P_{\hat \theta} \approx P_{\rm perturbed}^{(\rm new)}$,  where
\begin{eqnarray}
\label{theta hat implementation}
\hat \theta  =  {\rm arg} \min_{\theta \in {\cal S}_{I D}} \left \| L_{\hat \alpha} \hat M_Q^{({\rm new})} - P_{\theta} \right\|_{{\cal H}}^2, \quad {\cal S}_{I D} = \{x \in [0, 1]^{I D }, \ 1^T x = 1  \}.
\end{eqnarray}
\end{step}

\subsection{Modelling perturbations in the London Underground}
\label{section modelling the tube}
\setcounter{step}{0}

The predictive power of the distribution regression interference model outlined in 
Section \ref{section implementation} is
tested on a real-world data consisting of observed exit counts from
the London underground.  We consider a graph with $D = 269$ nodes
corresponding to the stations in the whole London Underground (all 11
underground lines are included) and adjacency matrix defined by
\begin{eqnarray}
A_{dd'} = \left\{ \begin{array}{ll}
1 & $if there exists a link between station $ d $ and station $ d'\\
0 & $otherwise$ 
\end{array} \right. , \quad d, d'=1, \dots D.
\end{eqnarray}
The natural regime dataset, ${\cal D}_{\rm natural}$, consists of $N = 35$ days of input-output records from late 2013
\begin{eqnarray}
\label{normal regime datasets}
 \{ {\cal D}_{\rm day} \}_{{\rm day} = 1}^{N},
\qquad {\cal D}_{\rm day} = \{u^{(m)} = [o, d, t_o, t_d], o, \ d \in \{ d'\}_{d'=1}^D, \ t_o, t_d \in T  \}_{m = 1}^{M_{\rm day}},  
\end{eqnarray}
where $M_{\rm day}$ (${\rm day} = 1, \dots, N$) is the number of journey records for ${\rm day}$,  $T = \{ T_{\min}, T_{\min} +1, \dots, T_{\max}\} $ the
`observation' time window, $o$ and  $d$ the origin and destination
journey $u^{(m)}$, and $t_o$ and $t_d$ the corresponding starting and exit
times.  
We have access to the features of $K = 72$ observed disruptions 
\begin{eqnarray}
{\cal D}_{\rm disruptions}  = \{ z^{(k)} = [{\rm day}, t_{\rm start}, t_{\rm end}, {\rm ROI}], {\rm day} \in \{n\}_{n=1}^N,  t_{\rm start}, t_{\rm end} \in T, {\rm ROI} \subset \mathcal V \}_{k=1}^K, 
\end{eqnarray}
where ${\rm day}$ is the disruption's day, $t_{\rm start}, t_{\rm
	end}$ the disruption's starting and ending times and ${\rm ROI}$ (``region of interest'')
the list of stations that are endpoints of some disrupted link.

For each day and each minute in a day,  we compute the number of journeys
completed between station $o  \in \{ d'\}_{d'=1}^D$ and station $d \in  \{ d'\}_{d'=1}^D$
\begin{eqnarray}
\label{realizations normal regime}
\left[ y^{({\rm day})}\right]_{odt} = \sum_{u \in {\cal D}_{\rm
		day}} \delta_{u_1, o} \delta_{u_2, d} \delta_{u_4, t}, \quad o,
d = \{ d'\}_{d'=1}^D, \quad t \in T\quad {\rm day} \in \{n\}_{n=1}^N, 
\end{eqnarray}
which can be interpreted as $N - 1$ exchangeable samples of a natural
regime random variable $Y \sim P_{\rm natural}$ (for each disruption,
we exclude the sample associated with the disruption day).  For each
disruption $[z_1, z_2, z_3, {\rm ROI}] = z^{(k)} \in {\cal D}_{\rm disruptions}$, we want to describe a random
variable $\tilde Y  \in {\mathbf N}_+^{|{\rm ROI}|} $,  $\tilde Y \sim P_{\rm perturbed}^{(k)}$ associated with the number of exits in region of
interest. 
Here we have access to a single realization, $\tilde y^{(k)}  \in {\mathbf N}_+^{|{\rm ROI}|}$, of $\tilde Y$,  which is defined by 
\begin{eqnarray}
[\tilde y^{(k)}]_j =  \sum_{o = 1}^D \sum_{t = z_2}^{z_3} [y^{(z_1)}]_{o d_j t}, \quad d_j = {\rm ROI}_j, \quad j=1, \dots, |{\rm ROI}|, 
\end{eqnarray} 
where $y^{({\rm day})}$ is defined in \eqref{realizations normal regime}.
We let the input variables associated with the $k$-th output, $P_{\rm perturbed}^{(k)}$, be the probability distributions, $\{ Q_i^{(k)} \}_{i = 1}^I$, associated with a set of random variables $\{ X^{(k)}_i \in {\mathbf N}_+^{|{\rm ROI}|}, X^{(k)}_i = f^{(k)}_i(Y), Y \sim P_{\rm natural}  \}_{i = 1}^I$,  where $\{f^{(k)}_i\}_{i = 1}^I$ are deterministic function of $Y \sim P_{\rm natural}$.
In particular, we choose $I = 5$ and  $\{f^{(k)}_i\}_{i = 1}^I$ defined by 
\begin{align}
\label{input variables tube}
&\left[X^{(k)}_1 \right]_j = \sum_{o = 1}^D \sum_{t = z_2}^{z_3}  {\bf 1}_{g(o, d_j, A, \tilde A) \leq \xi} Y_{o d_j t}, \quad 
\left[X^{(k)}_2 \right]_j = \sum_{o = 1}^D \sum_{t = z_2}^{z_3}  {\bf 1}_{g(o, d_j, A, \tilde A)  > \xi} Y_{o d_j t}, \\ 
&\left[X^{(k)}_3 \right]_j =\sum_{o = 1}^D  \sum_{t = z_2}^{z_3} Y_{o d_j t}, \quad 
\left[X^{(k)}_4 \right]_j = \sum_{o = 1}^D  \sum_{t = z_2}^{z_3} E(Y_{o d_j t}), \quad
X^{(k)}_5  =  \sum_{d \in {\rm ROI}} \sum_{o = 1}^D  \sum_{t = z_2}^{z_3} Y_{o d t}  1^{(k)},  \nonumber 
\end{align}
where $d_j = {\rm ROI}_j$, $j=1, \dots, |{\rm ROI}|$, ${\bf 1}_{u} = 1$ if $u$ is verified and zero otherwise, $g(o, d, A, A') = 1 - {\rm dist}_{A'}(o, d) / {\rm dist}_{A}(o, d)$ with  ${\rm dist}_A(d, d')$ being the distance between $d$ and $d'$ computed on the graph associated with adjacency matrix $A$, $\tilde A^{(k)}$ the adjacency matrix corresponding to disruption $z^{(k)} \in {\cal D}_{\rm disruptions}$ defined by
\begin{eqnarray}
\left[ \tilde A^{(k)}\right]_{dd'} = \left\{ \begin{array}{cc} 
0 & $ if $ d \in {\rm ROI} $ or $ d' \in {\rm ROI} \\
A_{dd'} & $otherwise$  
\end{array}  \right. ,
\end{eqnarray}
$\xi > 0 $ a threshold and $[1^{(k)}]_j = 1/ |{\rm ROI}|$ for all $j=1, \dots, |{\rm ROI}|$.
In terms of the new variables, the network interference problem presented in Section \ref{sec:setup} is reformulated 
as a standard  RKHS distribution-to-distribution regression of the type analyzed in Section \ref{sec:method}
\begin{eqnarray}
\Psi(P_{\rm natural}, z^{(k)}, A)  =  \Psi(\{ Q_{i}^{(k)} \}_{i = 1}^I).
\end{eqnarray}
The estimation of the regression model and the approximate sampling from a predicted perturbed distribution $P_{\rm perturbed}^{({\rm new})}$ are obtained as in Section \ref{sec:method} and \ref{section implementation}. 
In particular, as the support of the output distributions depends on the features of the new disruption, $z^{({\rm new})}$, \emph{i.e.}, $P_{\rm perturbed}^{({\rm new})}: {\mathbf N}_+^{|{\rm ROI}|} \to [0, 1]$ ($[z_1, z_2, z_3, {\rm ROI}] = z^{({\rm new})}$), the following features-specific basis is constructed from a set of \emph{rescaled marginals} of $P_{\rm natural}$: 
\begin{eqnarray}
\{ \{ U_{rj}  \in {\cal P}^{({\rm new})} \ {\rm \ s.t. \ } X_{rj} \sim U_{rj}, \ X_{rj} = \lambda_{r} \sum_{o = 1}^D \sum_{t = z_2}^{z_3}
Y_{od_jt} \}_{j= 1}^{ |{\rm ROI}|} \}_{r = 1}^R, 
\end{eqnarray}  
where $[z_1, z_2, z_3, {\rm ROI}] = z^{({\rm new})}$, $d_j = [{\rm ROI}]_{j}$, $j=1, \dots, |{\rm ROI}|$, $Y \sim P_{\rm natural}$, 
$\lambda_{r} = 1 + (r - 1) C$, $C = c(R - 1)^{-1} \max_j E_{Y \sim P_{\rm natural}}(\sum_{o = 1}^D \sum_{t = z_2}^{z_3} Y_{od_jt})$, $ c > 1$ and  $R \in {\mathbf N}_+$.

\subsection{Empirical results}
\label{section empirical results}

\begin{figure}
	\begin{center}
		\includegraphics[width = .5\textwidth]{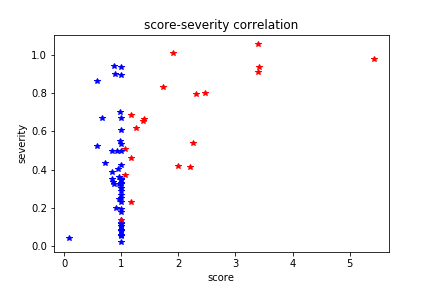}
	\caption{Correlation between the unobservable severity defined in \eqref{severity} (y-axis) and its observable proxy defined in \eqref{observable score} (x-axis).  
		Stars correspond to disruptions in the original dataset with red markers highlighting those that have been used in the experiment.
	}
	\label{figure correlation severity}
	\end{center}
\end{figure}

\begin{figure}
	\begin{center}
		\includegraphics[width=.49\textwidth]{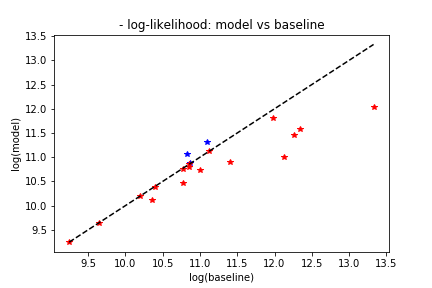} 
		\includegraphics[width=.49\textwidth]{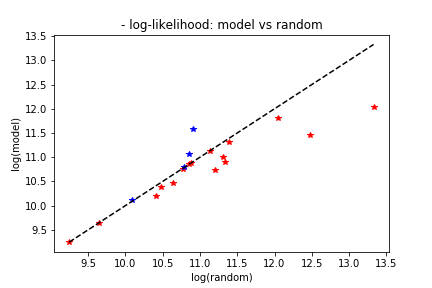} 	
	\end{center}	
		\caption{Logarithm of the negative log-likelihood of the disruption-day exit counts, \emph{i.e.}, $\log( - \log(p(\hat y)))$ for $\hat y \in {\cal D}_{P_{\rm model}^{(k)}}$, with $P^{(k)}_{\rm model}$ being the baseline model (left, x-axis), a random model (right, x-axis) and the proposed model (left and right, y-axis).
		Red markers highlight cases where the proposed model outperforms the models to which it is compared. 
	}
	\label{figure score ll}
\end{figure}

\begin{figure}
	\begin{center}
		\includegraphics[width=.49\textwidth]{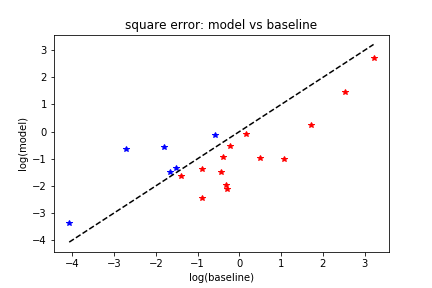} 
		\includegraphics[width=.49\textwidth]{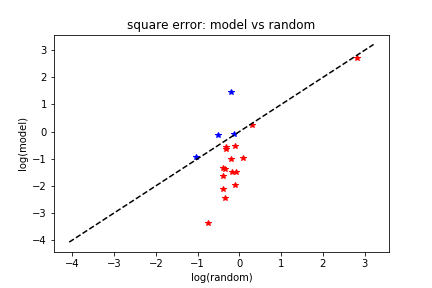} 		
	\end{center}
	\caption{Logarithm of the square error $\|E_{Y \sim P^{(k)}_{\rm model}}(Y) - \hat y \|^2\|\hat y\|^{-2}$,  with $\hat y \in {\cal D}_{P_{\rm perturbed}^{(k)}}$ and $P^{(k)}_{\rm model}$ being the baseline model (left, x-axis), a random model (right, x-axis) and the proposed model (left and right, y-axis).
		Red markers highlight cases where the proposed model outperforms the models to which it is compared.
	}
	\label{figure score se}
\end{figure}

\begin{figure}		
	\begin{center}
		\foreach \index in {0, ..., 8} 
		{
			\includegraphics[width=.32\textwidth]{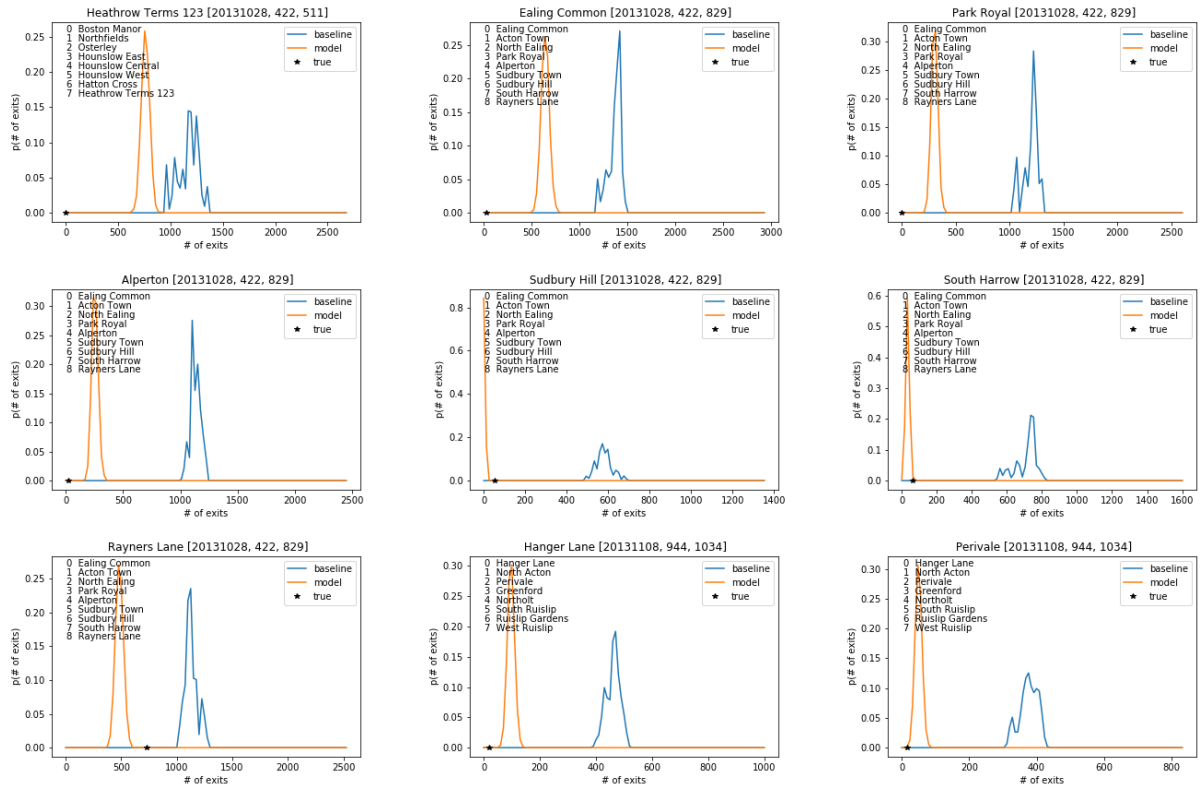}
		}
	\end{center}
	\caption{Exit counts probability distributions (selection 1): disruptions and stations where the baseline model obtained the lowest likelihood (per station).}
	\label{figure distributions 1}	
\end{figure}

\begin{figure}		
	\begin{center}
		\foreach \index in {0, ..., 8} 
		{
			\includegraphics[width=.32\textwidth]{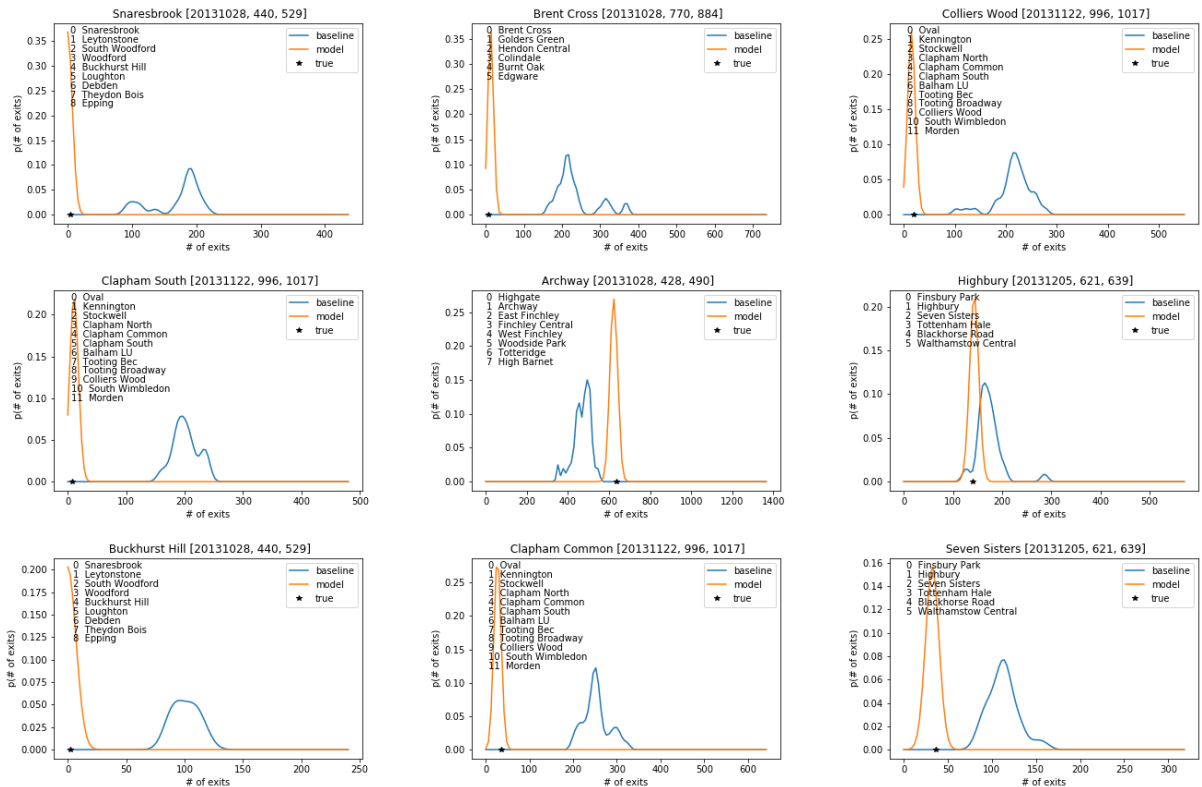}
		}
	\end{center}
	\caption{Exit counts probability distributions (selection 2): disruptions stations where the proposed model obtained the highest likelihood (per station).}
	\label{figure distributions 2}
\end{figure}

To test the London Underground model outlined in Section \ref{section modelling the tube}, we have created a reduced dataset of disruptions by selecting the $N = 20$ disruptions with highest \emph{observable score}  
\begin{eqnarray}
\label{observable score}
{\rm score}^{(k)} = \frac{\sum_{{\rm day}= 1}^{N} {\bf 1}_{{\rm day} \neq z_1} \| x_1^{({\rm day})} -x_2^{({\rm day})} \|^2}{ \sum_{{\rm day}= 1}^{N} {\bf 1}_{{\rm day} \neq z_1} \| x_1^{({\rm day})} \|^2},  \quad x_i^{({\rm day})} \in {\cal D}_{Q_i^{(k)}}, \quad  i= 1, 2 , 
\end{eqnarray}
where $[z_1, z_2, z_3, {\rm ROI}] = z^{(k)} \in {\cal D}_{\rm disruptions}$, $\{ Q_i^{(k)} \}_{i=1, 2}$ are the distributions describing the input random variables $\{ X_i^{(k)} \}_{i=1, 2}$ defined in \eqref{input variables tube} and $\{ {\cal D}_{Q_i^{(k)}} \}_{i=1, 2}$ the corresponding realizations sets obtained from the normal regime datasets defined in \eqref{normal regime datasets}.
The observable score for disruption $k$ depends on the difference between $x_1^{({\rm day})}$ (${\rm day} = 1, \dots, N$, ${\rm day} \neq z_1$), the number of people exiting at $d \in {\rm ROI}^{(k)}$ from paths that are \emph{feasible} on the disruption day, \emph{i.e.}, for $z_1 = z^{(k)}$, and $x_2^{({\rm day})}$ (${\rm day} = 1, \dots, N$, ${\rm day} \neq z_1)$, and the number of people exiting at $d \in {\rm ROI}$ from paths that are \emph{infeasible} on the disruption day. 
We refer to this score as `observable' because it can be computed before measuring the effects of a disruption, given the disruption's features $z^{(k)} \in {\cal D}_{\rm disruptions}$ and the natural regime datasets \eqref{normal regime datasets}.
The score is a proxy for the `unobservable' \emph{severity score}   
\begin{eqnarray}
\label{severity}
{\rm severity}^{(k)} = \frac{\sum_{j = 1}^{|{\rm ROI}^{(k)}|} (\sum_{o = 1}^D \sum_{t = z_2}^{z_3} (
E(Y_{od_jt} )- y^{(z_1)}_{od_jt}))^2}{ \sum_{j = 1}^{|{\rm ROI}^{(k)}|} (\sum_{o = 1}^D \sum_{t = z_2}^{z_3} 
E(Y_{od_jt}))^2}, 
\end{eqnarray}
where $[z_1, z_2, z_3, {\rm ROI}] = z^{(k)} \in {\cal D}_{\rm disruptions}$, $d_j = [{\rm ROI}^{(k)}]_{j}$, $y^{(z_1)} \in {\cal D}^{(z_1)}$ defined in \eqref{normal regime datasets}, $Y \sim P_{\rm natural}$.
The severity score is unobservable because it is not available before observing the effects of a disruption.
Figure \ref{figure correlation severity} shows the approximate correlation between observable scores and true severity scores, with disruptions selected for the experiment marked in red.
The model evaluation has been performed by splitting the dataset of selected disruptions in $10$ subsets and running $10$ training-evaluation instances.
Each instance corresponds to a different test set, with the model trained on the remaining nine subsets.
Thus, all averages and log-likelihood evaluations reported in Figures \ref{figure score ll}, \ref{figure score se}, \ref{figure distributions 1} and \ref{figure distributions 2} are out-of-sample predictions, with models tested on disruptions that were not used for training. 
For computational efficiency, the value of the kernel parameter $\rho$ was obtained through cross-validation over the all sample 
and kept fixed over all training-testing instances.
The usual procedure would be to fix $\rho$ by cross-validation on the training set of each training-testing instance. 
Figure \ref{figure score ll} and \ref{figure score se} show the predictive performance of the proposed model against a \emph{baseline} model and a set of \emph{random} models.
The baseline model is the empirical distribution obtained from the natural regime dataset.
The random models are obtained by letting $\hat \theta$ be a realization of $\Theta \sim {\rm uniform}(I |{\rm ROI}^{(k)}|)$ in the definition of the approximating mixtures needed for sampling (with the same basis elements used for the proposed model).
Figure \ref{figure distributions 1} and \ref{figure distributions 2} compare the shapes of station-specific densities associated with the proposed model and the baseline (densities associated with the random models are all similar and removed for visual reasons.). 
Predictions and log-likelihood evaluations are all (pseudo-)empirical estimations obtained by sampling from the models according to the sampling procedure described in Section \ref{section sampling procedure}.
For the baseline model we used are the true natural regime's samples.
Let $P^{(k)}_{\rm model} \in \{P^{(k)}_{\rm opt}, P^{(k)}_{\rm random}, P^{(k)}_{\rm baseline} \} $,  where 
$P^{(k)}_{\rm opt} = P_{\hat \theta}^{(k)}$ is the output of the proposed regression model with $P_{\hat \theta}^{(k)}$ being the approximating mixture described in Section \ref{section implementation} and $\hat \theta$ the estimated mixing weights vector \eqref{theta hat implementation}, 
$P^{(k)}_{\rm random} =  P_{\tilde \theta}^{(k)}$ is the random model mentioned above with $ P_{\tilde \theta}^{(k)}$ being the approximating mixture described in Section \ref{section implementation} and  random weights $\tilde \theta$ and 
$P^{(k)}_{\rm baseline}$ is the baseline model defined implicitly by $\tilde Y \sim P^{(k)}_{\rm baseline}$, $\tilde Y  =  \sum_{o = 1}^D \sum_{t = z_2}^{z_3} Y_{od_jt}$ ($Y \sim P_{\rm natural}$, $[z_1, z_2, z_3, {\rm ROI}] = z^{(k)} \in {\cal D}_{\rm disruptions}$, $d_j = [{\rm ROI}^{(k)}]_{j}$, $y^{(z_1)} \in {\cal D}^{(z_1)}$).
For each $k=1, \dots, K$, all models output a multivariate distribution, $P^{(k)}_{\rm model}$, describing a vector-valued random variable $\tilde Y \in {\mathbf N}_+^{|{\rm ROI}^{(k)}|}$. 
The station-specific marginal densities 
$p_{\rm model}^{(k)}(\tilde y_j)  = \left(\prod_{j' \neq j} \sum_{\tilde y_{j'} = 1}^{\rm \infty} \right) p_{\rm model}^{(k)}(\tilde y_1, \dots, \tilde y_{|{\rm ROI}^{(k)}|})$ ($j =1, \dots, |{\rm ROI}^{(k)}|$, $k= 1, \dots, K$) shown in Figure \ref{figure distributions 1} and \ref{figure distributions 2} are computed through the entry-wise density estimators
\begin{eqnarray}
p_{\rm model}^{(k)}(y_j) \propto \sum_{\tilde y \in {\cal D}_{U}} e^{ - h (y_j -  y'_j )^2},   \quad j =1, \dots, |{\rm ROI}^{(k)}|, \quad U = P_{\rm model}^{(k)}, 
\end{eqnarray}
where ${\cal D}_{U}$ is a set of realizations of $\tilde Y \sim U$ ($U \in \{ P^{(k)}_{\rm model}\}_{k=1}^K $) and  $h > 0$ is a smoothing parameter.

%

\section*{Acknowledgements}

We thank Transport for London for kindly providing access to
data. This work was supported by a ESPRC grant EP/N020723/1 to NC, RS
and SMK. AG acknowledges funding from the Gatsby Charitable
Foundation. RS received further support from the Alan Turing
Institute, under the EPSRC grant EP/N510129/1.

\appendix

\section*{Appendix}
\subsection*{Proofs}

\begin{refproof}[of Lemma~\ref{lemma consistency non parametric}]
	Assume that the meta-distribution ${\mathbf P}: {\cal P} \times {\cal
		Q} \to [0, 1]$ introduced in \eqref{meta distribution} can be
	represented by a \emph{unique} RKHS linear operator ${\cal L}_{{\cal
			Q}{\cal P}}$ such that
	\begin{eqnarray}
	\mu_{P} = L_{\rm true} \mu_{Q},
	\end{eqnarray}
	for any possible pair of input-output distributions $[P, Q] \sim
	{\mathbf P}$.  For simplicity, we assume that \eqref{exact model non
		parametric} holds exactly but misspecification terms can also be
	added with minor changes to the following.  Assumption \eqref{exact
		model non parametric} implies directly
	\begin{eqnarray}
	\label{nu extra empirical}
	\hat \mu_{P^{(k)}} &=& L_{\rm true} \hat \mu_{Q^{(k)}} + \hat \nu_{\rm extra}^{(k)},  \qquad k = 1, \dots, K,
	\end{eqnarray}
	where the empirical mean embeddings $\{ \{\hat \mu_U^{(k)}\}_{k = 1}^K
	\}_{U \in \{P, Q\}}$ are associated with the dataset ${\cal D}^{(K)}$
	defined in \eqref{datasets}.  The error terms $\{ \hat \nu_{\rm extra}^{(k)} \}_{k = 1}^K$
	obey
	\begin{eqnarray}
	\label{nu extra bound}
	\| \hat \nu_{\rm extra}^{(k)} \|_{\cal H} &\leq& \| \hat \mu_{P^{(k)}} - \mu_{P^{(k)}}\|_{\cal H} + \| L_{\rm true} \| \| \hat \mu_{Q^{(k)}} - \mu_{Q^{(k)}}\|_{\cal H} = O\left(1/D^{(K)}\right), \nonumber \\
	D^{(K)} &=& \min\{ \min\{ |{\cal D}_{P^{(k)}}|, |{\cal D}_{Q^{(k)}}|\}, \{ {\cal D}_{P^{(k)}}, {\cal D}_{Q^{(k)}}\} \in {\cal D}^{(K)} \}.
	\end{eqnarray}
	where $\| L \| = \max_{\| v \|_{\cal H} = 1} \| L v \|_{\cal H} $.
	Equation  \eqref{nu extra empirical} implies 
	\begin{eqnarray}
	\hat M_P = L_{\rm true} \hat M_Q + \hat N_{\rm extra}, \qquad
	\hat N_{\rm extra}  = [\hat \nu^{(1)}_{\rm extra}, \dots, \hat \nu^{(K)}_{\rm extra}].
	\end{eqnarray}
	For any $K \in {\mathbf N}_+$, we define 
	\begin{eqnarray}
	\tilde {\cal H}_U  = \{ \hat \Pi_{U} f, \  f \in {\cal H}_{U} \}, \qquad
	\hat \Pi_{U}^{(K)}  = \hat M_U (\hat M_U^T \hat M_U)^{-1} \hat M_U^T, \qquad
	U \in \{P, Q\},
	\end{eqnarray}
	which are the subspaces of $\{ {\cal H}_{U} \}_{U \in \{P, Q \}} $
	spanned by the columns of $\{\hat M_U \}_{U \in \{P, Q \} }$.  We
	prove the consistency of the estimator \eqref{estimator} as a map
	between $\tilde {\cal H}_Q$ and $\tilde {\cal H}_Q$ as follows
	\begin{eqnarray}
	\Delta^{(K)} & = & \| \hat \Pi_{P}^{(K)} \left( L_{\rm true} - \hat L^{(K)} \right) \hat \Pi_{Q}^{(K)} \| \\
	& = & \| \hat \Pi_{P}^{(K)} \left(L_{\rm true} - \hat M_P (\hat M_Q^T \hat M_Q)^{-1} \hat M_Q^T  \right) \hat \Pi_{Q}^{(K)} \| \\
	& = & \| \hat \Pi_{P}^{(K)} \left(L_{\rm true} - (L_{\rm true} \hat M_Q + \hat N_{\rm extra}) (\hat M_Q^T \hat M_Q)^{-1} \hat M_Q^T  \right) \hat \Pi_{Q}^{(K)}   \| \\
	&\leq &\| \hat \Pi_{P}^{(K)} (L_{\rm true} - L_{\rm true} \hat \Pi_{Q}^{(K)}) \hat \Pi_{Q}^{(K)}   \| + \| \hat \Pi_{P}^{(K)} \hat N_{\rm extra} (\hat M_Q^T \hat M_Q)^{-1} \hat M_Q^T  \| \\
	&\leq &\| \hat \Pi_{P}^{(K)} L_{\rm true} \hat \Pi_{Q}^{(K)}  - \hat \Pi_{P}^{(K)}  L_{\rm true} \hat \Pi_{Q}^{(K)}  \| + \| \hat N_{\rm extra} (\hat M_Q^T \hat M_Q)^{-1} \hat M_Q^T \| \\
	&\leq & \sqrt{\sigma_{\rm max}(\hat N_{\rm extra}^T \hat N_{\rm extra}) \sigma_{\min}(\hat M_Q^T \hat M_Q)},
	\end{eqnarray}
	where 
	\begin{eqnarray}
	\sigma_{\rm max}(\hat N_{\rm extra}^T \hat N_{\rm extra}) \leq \sum_{k, k'} \| \nu^{(k)}_{\rm extra} \|_{\cal H}^2 \| \nu^{(k')}_{\rm extra} \|_{\cal H}^2 =  O\left(1/D^{(K)}\right)
	\end{eqnarray}
	This implies $\Delta^{(K)} = O(1/D^{(K)})$ and hence the statement.
\end{refproof}

\begin{refproof}[of Lemma~\ref{lemma consistency of the one-parameter model}]
	Assuming an exact model $\mu_{P^{(k)}} = \bar \alpha \mu_{Q^{(k)}}$ for all $k = 1, \dots, K$, implies
	\begin{eqnarray}
	\mu_{P^{(k)}}  =  \bar \alpha \hat \mu_{Q^{(k)}} + \hat \nu_{\rm extra}^{(k)}, \qquad k = 1, \dots, K,
	\end{eqnarray}
	where $\{ \hat \nu_{\rm extra}^{(k)} \}_{k = 1}^K$ obey  
	\begin{eqnarray}
	\|\hat \nu_{\rm extra}^{(k)} \|_{\cal H} \leq \|\hat \mu_{P^{(k)}} -
	\mu_{P^{(k)}} \|_{\cal H} + \bar \alpha \| \hat \mu_{Q^{(k)}} -
	\mu_{Q^{(k)}} \|_{\cal H} = O\left(1/ D^{(K)}\right),
	\end{eqnarray}
	with $D^{(K)} = \min\{ \min\{ |{\cal D}_{P^{(k)}}|, |{\cal D}_{Q^{(k)}}|\}$ ($ {\cal D}_{P^{(k)}}, {\cal D}_{Q^{(k)}} \in {\cal D}^{(K)} $).
	This implies 
	\begin{eqnarray}
	\hat M_P = \bar \alpha \hat M_Q + \hat N_{\rm extra}, \qquad \hat N_{\rm extra} =  [\hat \nu^{(1)}, \dots, \hat \nu^{(K)}],
	\end{eqnarray}
	and hence $\hat m_{QP} = \bar \alpha \hat m_{QQ} + \hat N_{\rm extra}^T \hat M_{Q}$,  
	where 
	\begin{eqnarray}
	\| \hat N_{\rm extra} \|_{\cal H} \leq \sum_{k = 1}^K \| \hat \nu^{(k)}_{\rm extra} \|_{\cal H} = O\left(1/D^{(K)}\right).
	\end{eqnarray}
	Then
	\begin{eqnarray}
	|\hat \alpha^{(K)}  - \bar \alpha |  &=& \left| {\rm trace}(\hat m_{QQ})^{-1} \left( {\rm trace}( \hat m_{PQ}) - \bar \alpha \  {\rm trace}(\hat m_{QQ})\right)   \right| \\ 
	&=& \left| {\rm trace}(\hat m_{QQ})^{-1}  {\rm trace}( \hat N_{\rm extra}^T \hat M_{Q})  \right| \\
	&\leq& \frac{1}{\sigma_{\min}(\hat m_{QQ})}  \| \hat N_{\rm extra} \|_{\cal H} \| \hat M_{Q}\|_{\cal H}  \\
	& = & O\left(1/ D^{(K)}\right).
	\end{eqnarray}
\end{refproof}

\begin{refproof}[of Lemma~\ref{lemma consistency of the mixture of embedding model}]
	When \eqref{mixture of embeddings} holds exactly, it implies  
	\begin{eqnarray}
	\hat \mu_{P^{(k)}} = \hat M_Q^{(k)} \bar \alpha + \hat \nu_{\rm extra}^{(k)}, \qquad k = 1, \dots, K,
	\end{eqnarray}
	with $\{\hat \nu_{\rm extra}^{(k)} \}_{k = 1}^K$ obeying 
	\begin{eqnarray}
	\| \hat \nu_{\rm extra}^{(k)}  \|_{\cal H}  &=& \| \hat \mu_{P^{(k)}} - \sum_{i = 1}^I \bar \alpha_i  \hat \mu_{Q_i^{(k)} }  \|_{\cal H} \\
	&\leq &\| \hat \mu_{P^{(k)}} - \mu_{P^{(k)}}\|_{\cal H}  + \sum_{i = 1}^I \bar \alpha_i \| \hat \mu_{Q_i^{(k)}} - \mu_{Q_i^{(k)}}\|_{\cal H} \\
	& = & O\left(1/D^{(K)}\right),
	\end{eqnarray}
	and $D^{(K)} = \min \{ |{\cal D}_{U^{(k)}}|  \}_{k = 1}^K$. This implies
	\begin{eqnarray}
	\label{bound mixture of embeddings}
	\| \bar \alpha  - \hat \alpha^{(K)} \| & = & \| \bar \alpha - (\hat m^{(K)} )^{-1}  \sum_{k = 1}^K (\hat M_Q^{(k)})^T \hat \mu_{P^{(k)}}  \| \\
	& = & \| \bar \alpha - (\hat m^{(K)} )^{-1}  M^{(K)} \bar \alpha +  (\hat m^{(K)} )^{-1}  \sum_{k = 1}^K (\hat M_Q^{(k)})^T \hat \nu_{\rm extra}^{(k)} \|\\
	&\leq & \frac{1}{\sigma_{\min}(\hat m^{(K)})}  \| \sum_{k = 1}^K (\hat M_Q^{(k)})^T \hat \nu_{\rm extra}^{(k)} \| \\
	&\leq & \frac{1}{\sigma_{\min}(\hat m^{(K)})} \sum_{k = 1}^K \sum_{i = 1}^I \langle \hat \mu_{Q_i^{(k)}} ,  \hat \nu_{\rm extra}^{(k)} \rangle_{\cal H}  \\
	&\leq & \frac{1}{\sigma_{\min}(\hat m^{(K)})} \sum_{k = 1}^K \sum_{i = 1}^I  \|  \hat \mu_{Q_i^{(k)}} \|_{\cal H}  \|  \hat \nu_{\rm extra}^{(k)} \|_{\cal H} \\ 
	& =& O\left(1/D^{(K)}\right).
	\end{eqnarray}
\end{refproof}

\begin{refproof}[of Lemma \ref{lemma consistency of the mixture of distributions model}]
	Let the meta-distribution ${\mathbf P}: {\cal P}^{\otimes
		I} \times {\cal P} \to [0, 1]$ generating the realizations dataset 
	${\cal D}^{(K)}$ be such that
	\begin{eqnarray}
	P = \sum_{i=1}^I Q_i \bar w_i , \quad \bar w \geq 0, \quad 1^T \bar w = 1, 
	\end{eqnarray}
	for any $[P, \{Q_i \}_{i = 1}^I] \sim {\mathbf P}$.
	Equation \eqref{hilbert operator mixture of distributions equation} and Definition \ref{definition mixture of distributions} imply that the meta-distribution ${\mathbf P}$ is also such that 
	\begin{eqnarray}
	\mu_P = L_{\Psi}[ \mu_{Q_1}, \dots, \mu_{Q_I}],  \quad  L_{\Psi} = \bar w \otimes 1_{\cal H}, \quad \bar w \geq 0, \quad 1^T \bar w = 1;
	\end{eqnarray}  
	for any $[P, \{Q_i \}_{i = 1}^I] \sim {\mathbf P}$.
	This means that $\bar w$ can be estimated by solving a RKHS linear least-squares problem analogous to the one defined in \eqref{estimator mixture of embeddings}.
	Moreover, as $L_{\Psi}$ belongs to a sub-class of the models defined in Definition \ref{definition mixture of embeddings}, the consistency of \eqref{estimator mixture of distributions} can be obtained directly from Lemma \ref{lemma consistency of the mixture of embedding model} by adding the extra constraint $w \in {\cal S}_I$, with ${\cal S}_I$ defined in \eqref{estimator mixture of distributions}. 
	Since $w \in {\cal S}_I$ is a convex constraint, the solution, $\hat w$, of \eqref{estimator mixture of distributions} is unique and, for any $K \in {\mathbf N}_{+}$, obeys
	\begin{eqnarray}
	\| \bar w - \hat w^{(K)}\| \leq \| \bar w - \hat \alpha^{(K)} \|  = O\left(1/D^{(K)}\right),
	\end{eqnarray}     
	where $\hat \alpha \in {\mathbf R}^{I}$ is the solution of the unconstrained problem \eqref{estimator mixture of embeddings} and the equality follows from Lemma \ref{lemma consistency of the mixture of embedding model}. 
\end{refproof}

\begin{refproof}[of Lemma \ref{lemma mean embedding sampling}]
	Let $\hat \mu$ be a given empirical mean embedding, $P_{\hat \mu} \in {\cal P}$  
	the unknown distribution associated with $\hat \mu$ and  
	$\{ P_i \in {\cal P} \}_{i = 1}^I$  a suitable
	finite-dimensional basis of the functions space ${\cal P}$.
	Then $P_{\hat \mu} \in {\cal P}$ can be approximated by a superposition of 
	$\{ P_i \in {\cal P} \}_{i = 1}^I$
	\begin{eqnarray}
	P_{\hat \mu}  = (1 - \epsilon) \sum_{i = 1}^I \bar \theta_i P_i + \epsilon P_{\epsilon},
	\quad \bar \theta, \epsilon > 0, \quad 1^T \bar \theta = 1,
	\end{eqnarray}
	where $\epsilon P_{\epsilon}$ denotes the part of $P_{\hat \mu}$ that
	is not captured by  $P_i$.
	From Section \ref{section mixture of distributions}, this implies 
	\begin{eqnarray} 
	\hat \mu = (1 - \epsilon) \sum_{i=1}^I \bar
	\theta_i \mu_{i} + \epsilon \mu_{\epsilon}, \qquad \bar \theta,
	\epsilon > 0, \qquad 1^T \bar \theta = 1.
	\end{eqnarray}
	where 
	$\{\mu_{i}\}_{i = 1}^I$ are the mean embeddings of $\{ P_i \in {\cal P} \}_{i = 1}^I$ and $\mu_{\epsilon}$ the mean embedding of the error term.
	Assume we can sample from the basis distributions $\{ P_i \in {\cal P} \}_{i = 1}^I$ and let 
	\begin{eqnarray}
	{\cal D}_i =  \{ x^{(n)} {\rm realization  \ of \ } X \sim P_i \}_{n = 1}^{N_i}, \qquad i = 1, \dots, 
	\end{eqnarray}
	be the corresponding realizations datasets.
	Then we can compute the associated empirical mean embeddings $\{ \hat \mu_i
	\in {\cal H} \}_{i = 1}^I$ as in \eqref{empirical mean embedding}   
	and obtain an estimate of the unknown mixture weights $\bar \theta$ from 
	\begin{eqnarray}
	\hat \theta = {\rm arg} \min_{\theta > 0, 1^T \theta = 1}  \| \hat \mu - \sum_{ i = 1}^I \theta_i \hat \mu_i \|_{\cal H}.
	\end{eqnarray}
	For small $\epsilon$ we have $P_{\hat \mu} \approx \sum_{i = 1}^I \bar \theta_i P_i \approx \sum_{i = 1}^I \hat \theta_i P_i$ and samples from $P_{\mu}$ can be obtained as realizations of a random variable $X$ described by the obtained mixture  
	\begin{eqnarray}
	X \sim \sum_{i=1}^I \hat \theta_i P_i.
	\end{eqnarray}
\end{refproof} 

\begin{refproof}[of Lemma \ref{lemma consistency of the sampling scheme}]
	Let $P_{\hat \theta}$ be the approximating mixture defined in Lemma \ref{lemma mean embedding sampling}.
	Let $\epsilon \in [0, 1]$ measure the discrepancy between the target distribution and the best possible approximating mixture $P_{\bar \theta}$ ($\theta \in {\cal S}_I$).
	Similarly, we can let $\epsilon \in [0, 1]$ be defined by 
	\begin{eqnarray}
	\label{approximating mixture consistency proof}
	P_{\hat \mu}  = (1 - \epsilon) \sum_{i = 1}^I \bar \theta_i P_i + \epsilon P_{\epsilon},
	\quad \bar \theta, \epsilon > 0, \quad 1^T \bar \theta = 1,
	\end{eqnarray}
	where $\bar \theta = {\rm arg} \min_{\theta \in {\cal S}_I} {\rm dist}(P_{\hat \mu}, \sum_{i = 1}^I \theta_i P_i)$ and $P_{\epsilon} \in {\cal P}$ is such that the equality in \eqref{approximating mixture consistency proof} holds. 
	We want to show that the difference between the empirical expectation of an arbitrary RKHS function, $f \in {\cal H}$, with respect with the target distribution, $P_{\hat \mu}$, and the empirical expectation of $f$ obtained from the sampling procedure described by Lemma \ref{lemma mean embedding sampling}  tends to zero when $\epsilon \to 0$, $|{\cal D}_{P_{\hat \mu}} | \to \infty$, $|{\cal D}_{P_{\hat \theta}} |\to \infty$, with ${\cal D}_{P_{\hat \mu}}$  and ${\cal D}_{P_{\hat \theta}}$ defined in Lemma \ref{lemma consistency of the sampling scheme}.
	From the definitions in Lemma \ref{lemma consistency of the sampling scheme} we have 
	\begin{eqnarray}
	\hat \Delta_f & = & \left| \frac{1}{| {\cal D}_{P_{\hat \mu}} | }  \sum_{x \in {\cal D}_{P_{\hat \mu}}} f(x)  - \frac{1}{|{\cal D}_{P_{\hat \theta}} |}  \sum_{x \in {\cal D}_{P_{\hat \theta}}} f(x) \right| \\ 
	& \leq & \left|E_{X \sim P_{\hat \mu}}(f(X))  - E_{X \sim P_{\hat \theta}}(f(X)) \right| + O\left(\frac{1}{|{\cal D}_{P_{\hat \mu}} |}\right) + O\left(\frac{1}{|{\cal D}_{P_{\hat \theta}} |}\right) \\
	& = & \left| \langle f, \hat \mu   - \mu_{P_{\hat \theta}} \rangle  \right|+ O\left(\frac{1}{|{\cal D}_{P_{\hat \mu}} |}\right) + O\left(\frac{1}{|{\cal D}_{P_{\hat \theta}}}\right) \\
	& \leq & \|   f \|_{\cal H}  \  \| \hat \mu   - \mu_{P_{\hat \theta}} \|_{\cal H}+ O\left(\frac{1}{|{\cal D}_{P_{\hat \mu}} |}\right) + O\left(\frac{1}{|{\cal D}_{P_{\hat \theta}}}\right) \\
	& = & O\left(\epsilon \right) + O\left(\frac{1}{|{\cal D}_{P_{\hat \mu}} |}\right) + O\left(\frac{1}{|{\cal D}_{P_{\hat \theta}}}\right).
	\end{eqnarray}
\end{refproof}

\bibliographystyle{plainnat}
\bibliography{20190820arxiv}

\end{document}